\documentclass[sigconf]{acmart}

\AtBeginDocument{%
  \providecommand\BibTeX{{%
    \normalfont B\kern-0.5em{\scshape i\kern-0.25em b}\kern-0.8em\TeX}}}

\copyrightyear{2022}
\acmYear{2022}
\setcopyright{acmlicensed}\acmConference[NLPIR 2022]{2022 6th International Conference on Natural Language Processing and Information Retrieval}{December 16--18, 2022}{Bangkok, Thailand}
\acmBooktitle{2022 6th International Conference on Natural Language Processing and Information Retrieval (NLPIR 2022), December 16--18, 2022, Bangkok, Thailand}
\acmPrice{15.00}
\acmDOI{10.1145/3582768.3582789}
\acmISBN{978-1-4503-9762-9/22/12}

\usepackage{algorithm} 
\usepackage{algpseudocode} 
\usepackage{csquotes}
\usepackage{caption}
\usepackage{subcaption}
\usepackage{xcolor, soul}
\sethlcolor{yellow}

\usepackage{appendix}
\usepackage{tabularx}
\usepackage{booktabs}
\usepackage{amsmath,amsfonts}
\usepackage{amsthm}
\usepackage{hyperref}

\usepackage{float} 
\hypersetup{
  colorlinks=true,
  citecolor=Violet,
  linkcolor=Red,
  urlcolor=Blue
  citebordercolor=Violet}

\emergencystretch=\maxdimen
\begin{document}

\title{A Semantic Approach to Negation Detection and Word Disambiguation with Natural Language Processing}


\author{Izunna Okpala}
\email{okpalaiu@mail.uc.edu}
\affiliation{
  \institution{University of Cincinnati}
  \streetaddress{P.O. Box 210123}
  \city{Cincinnati}
  \state{Ohio}
  \postcode{45220}
  \country{USA}
}
\author{Guillermo Romera Rodriguez}
\email{gkr5144@psu.edu}
\affiliation{
  \institution{Penn State University}
  \streetaddress{P.O. Box 3000}
  \city{Philadelphia}
  \state{Pennsylvania}
  \postcode{16801}
  \country{USA}
}
\author{Andrea Tapia}
\email{axh50@psu.edu}
\affiliation{
  \institution{Penn State University}
  \streetaddress{P.O. Box 3000}
  \city{Philadelphia}
  \state{Pennsylvania}
  \postcode{16801}
  \country{USA}
}

\author{Shane Halse}
\email{halsese@ucmail.uc.edu}
\affiliation{
  \institution{University of Cincinnati}
  \streetaddress{P.O. Box 210123}
  \city{Cincinnati}
  \state{Ohio}
  \postcode{45220}
  \country{USA}
}

\author{Jess Kropczynski}
\email{kropczjn@ucmail.uc.edu}
\affiliation{
  \institution{University of Cincinnati}
  \streetaddress{P.O. Box 210123}
  \city{Cincinnati}
  \state{Ohio}
  \postcode{45220}
  \country{USA}
}

\renewcommand{\shortauthors}{Okpala, et al.}

\begin{abstract}
This study aims to demonstrate the methods for detecting negations in a sentence by uniquely evaluating the lexical structure of the text via word-sense disambiguation. The proposed framework examines all the unique features in the various expressions within a text to resolve the contextual usage of all tokens and decipher the effect of negation on sentiment analysis. The application of popular expression detectors skips this important step, thereby neglecting the root words caught in the web of negation and making text classification difficult for machine learning and sentiment analysis. This study adopts the Natural Language Processing (NLP) approach to discover and antonimize words that were negated for better accuracy in text classification using a knowledge base provided by an NLP library called WordHoard. Early results show that our initial analysis improved on traditional sentiment analysis, which sometimes neglects negations or assigns an inverse polarity score. The SentiWordNet analyzer was improved by 35\%, the Vader analyzer by 20\% and the TextBlob by 6\%.
\end{abstract}




\maketitle

\section{Introduction}

Negation is an inescapable variable when it comes to human communication. To convey spoken words or written texts, negations need not to be neglected because they are integral to everyday expression \cite{coombs2015value,ifeanyi2014text}. Popular text mining techniques have been studied over the years as to how they can automatically detect emotions, attitudes, sentiments, and perceptions, to name a few \cite{jiao2019proposal}. Some notable features that have received less attention in text analysis are negations and multiple negatives \cite{baker1969double}. As a result, the overall accuracy of popular sentiment analyzers has decreased. Understanding negation cues is essential for looping every word in a sentence together and capturing the elements that were affected. 

The term "negation" is a linguistic feature that negates the meaning of its closest neighbor in a sentence; i.e., it is the semantic opposite of a phenomenon \cite{widdows2003orthogonal}. This feature (negation) often poses a challenge to text mining techniques, leading to incorrect classification. This is because, generally, text mining techniques and sentiment analysis programs classify texts using a combined factor of words that are commonly associated with emotions \cite{udebuana2019analysis}. However, most techniques or algorithms never factor in the effect of negations on the cluster of words. They only concentrate on the evaluation of the polarity score for a given sentence, thus creating a potential scenario for misclassification if the sentence contains negations. The typical way the computer handles negation is to invert the polarity of the lexical item that lies next to the negator in a phrase \cite{jurek2015improved} (e.g., good: 0.5 and not good: -0.5). In this study, we propose a different method. Rather than just inverting the value, we recommended developing a function that computes the polarity of the negated word based on a list of antonyms from 5 different dictionaries, as well as the overall mean score of the polarity of all the antonyms. 

This study explores a novel approach to detecting and analyzing negations within sentences. This was achieved using a mixed approach: the detection of negation and word sense disambiguation. The detection of negation occurs through the use of pre-defined labels representing the negation signals, while the processing is applied through the use of sequence labeling and an antonymization function that averages out the polarity of all the antonyms, putting Parts of Speech (POS) tagging into perspective. Early results showed that our approach improved on conventional sentiment analysis, which sometimes ignores word negations or assigns an inverse polarity score without considering the context of the negated words. The SentiWordNet analyzer was enhanced by 35\%, the Vader analyzer by 20\%, and the TextBlob analyzer by 6\%. 

\section{Background}

Negations are extremely important in all human languages. The highly complex but subtle expression of negation in natural language, as conveyed in parts of speech, defies the basic syntactic structure of logical negation (adverbs, verbs, adjectives, quantifiers) \cite{sepnegation}. Negations in literature are words that systemically negate another expression. Double negatives, on the other hand, a terminology often misinterpreted to mean negations, are two negative words occurring concurrently in a sentence \cite{widdows2003orthogonal,ovalle2004double}. For example, if we have the text, \textit{"the prisoner was a dreaded terrorist",} The two words \textit{"dreaded"} and \textit{"terrorists"} occurring in the same sentence are referred to as \textit{double negatives or multiple negatives.} An example of negation, however, is when we have \textit{"The warden is not good"}; the word \textit{"not"} negates the phrase, \textit{"good."} To shorten the sentence \textit{"the warden is not good,"} we have, \textit{"the warden is bad."} This form of the sentence structure contains a negation clause "not" and a positive adjective "good." For better interpretation by a non-native speaker or even a machine, the negating factor needs to cancel out the adjective and derive an alternative word \cite{ovalle2004double} i.e., when a negation and a positive are used together, the sentence is usually transformed into a negative one, vice versa; mathematically we would have \( -x * -x = x^2 \). The square of x \((x^2)\) shows that there are some permutations that took place after the removal of the negation sign.

In some instances where negations are not inverted, i.e., whenever the relational expression is not a phrase that modifies or qualifies an adjective, verb, or other adverb, or a phrase naming an attribute, attached to or grammatically related to a noun to change or describe it, the negation stands. A good example is in the text \textit{"Soldiers could \textbf{never attend} a parade."} The negation signal \textit{"never"} could not cancel out the word \textit{"attend"} because it does not modify or qualify an adjective, nor does it name an attribute. But a negation signal can negate a positive clause that modifies or qualifies and adjective; e.g., \textit{"The boy is not dirty"} becomes \textit{"The boy is clean"}. Mathematically, we represent this as: \( -x * x = -x^2 \).  Having negation in a sequence seems proper, but this burdens text classification and can be misleading \cite{kimberly2021double}. Nonetheless, they are occasionally used in everyday casual conversation, and numerous examples can be found on popular social media platforms \cite{quilty2019university}.

\subsection{Previous Work}

Research on negation detection is critical for building intelligent systems that use texts as the foundation of their computational process. Academics have studied the detection of negatives and negations. Some of the approaches used are rule-based, machine learning, and conditional random field (CRF) methods \cite{khandelwal2019negbert}. What these have in common is that they methodically addressed the subject of negation, which can be considered the foundation of this research. Rule-based techniques handle negation detection using static, non-dynamic rules \cite{mutalik2001use, chapman2001simple}. Machine learning methods focus on teaching a machine how to distinguish negations. It detects negations by automatically generating regular expression patterns \cite{rokach2008negation}, while the CRF follows the natural order of events in a sentence \cite{agarwal2010biomedical}.

\subsection{The Theory of Double Negatives}

The concept of words with negative connotation in languages is the foundation of double negatives \cite{jiaxuan2010division}. In standard English, each subject-predicate composition should only have one negative form for ease of interpretation, but double negatives are unavoidable due to the unconventional statement structure used in social discussions \cite{quilty2019university}. Though it's common to believe that double negatives are just an artificial deviation from the norm, in some languages, like Spanish, anything other than a double negative is grammatically improper, i.e., they utilize double negatives as a proper expression \cite{rio2015comparative}. This is not so in English. However, the use of double negatives is prevalent in casual forums, discussion boards, and social media platforms \cite{quilty2019university}. Often times, data is collected from those mediums for predictive analytics.

\subsection{Lexicon-Based Analysis for Dealing with Negations}
In order to understand the meaningful bits in a text, lexicon-based analysis can be used for negative context detection. This technique employs about 6300 words \cite{jurek2015improved}, and was constructed manually using SentiWordNet (a lexical resource for text mining) \cite{baccianella2010sentiwordnet} as the baseline. Each word in the lexicon has a rating ranging from 100 (the most negative) to 100 (the most positive). This study also found evidence that some words appear to have a neutral rating when handled alone, but when put in between other words, the rating changes. This is based on the construction or lexical structure of the sentence. Depending on the sentiment analysis tool to be used, this rating differs; some use a ratio of from -1 to +1 \cite{baccianella2010sentiwordnet,hutto2014vader}, while others use 0 to +1 rating \cite{baccianella2010sentiwordnet}. Those ratings denote the score given to each word in their data dictionary. The lexicon-based analysis adopts a probabilistic model. To make this work, a labelled dataset of  over a million data points was used \cite{jurek2015improved}. As demonstrated by \cite{jurek2015improved}, the approach for calculating the probabilities follows a recursive mechanism by going through the words represented in the sentence.

\subsection{Natural Language Processing (NLP)}

The growing complexity of text data has made NLP applications increasingly vital in analyzing large volumes of data \cite{gardner2018allennlp,okpala2022perception}. NLP solutions are one of the backbones of some intelligent models like BERT, XLNet, and GPT, which have advanced the cause of sentiment analysis \cite{topal2021exploring, kovaleva2021bert}, and machine language translation \cite{phadke2017multilingual}.

NLP-enabled models propel the understanding of language structure and interpretation. The role of NLP, therefore, is to autonomously interpret human languages to return contextual patterns \cite{verma2011natural}. This technology involves methods for extracting meaning from text data \cite{tixier2016automated}. Another notable application of NLP in the research space is in the monitoring of fake news or the detection of spam emails, which has proven to be successful \cite{manzoor2019fake}. The applicability of NLP comes in a variety of forms. In our everyday discourse, such as social media conversations, negation signals are employed. In practice, some NLP techniques used in this space are Word2Vec and GloVe \cite{yu2017refining}. BERT and XLNet deliver state-of-the-art outcomes as well \cite{devlin2018bert}. However, each model has its own set of constraints. Most algorithms underperform when confronted with negation, which is why it is critical to identify novel approaches to overcome this issue in the NLP domain.

\subsection{Word Negation and Sequence Labeling}

The accuracy of NLP models such as sentiment or perception analysis depends on word negation and sequence labeling. Some NLP models typically work by analyzing each word, or sequence of words, independently. These algorithms deconstruct text into its minimum units called tokens \cite{webster1992tokenization}. Once these tokens have been cleaned up, the algorithms read each word or tokenized entity independently and come up with a result based on the different tokens found in the sentence or text. However, one of the limitations of most sentiment analysis algorithms is that they do not make sense out of multiple words, but just one word at a time \cite{abirami2017survey}. This can create a problem where sentences are mislabeled or miscategorized, causing problems down the line \cite{fu2018lexicon}. For example, a sentence that reads \textit{"Vaccines are \textbf{not} bad"} could be mislabeled because the presence of the word \textit{not}, but we know that sentence does not carry a negative sentiment. 

While popular opinion mining tools like sentiment analyzers try to find the negatives and positives, it is important to make a clear distinction between the terms "negative" and "negation." The most common negation signals are \textit{not, non, never, and neither} \cite{Liang2020May}. Since these negation signals are considered to be stopwords in the Natural Language Toolkit (NLTK), spacy and sklearn \cite{Liang2020May}, it's crucial to understand how they affect NLP tasks. There have been different attempts to overcome this drawback. The most recent and state-of-the-art approach was Bidirectional Encoder Representations from Transformers (BERT). Unlike other approaches, the novelty of BERT resides in its bidirectional training capacity, which means that sentences are read in both directions to get the complete context of the sentence. The trick here is also that tokens are read at the same time and not in order (either left-to-right or right-to-left). This allows BERT to take the context of each word into account. Similar in its construction, XLNet \cite{topal2021exploring} improved its masking mechanism with peculiar assumptions during its pre-training stage, and improved over the work done by BERT. Despite these advances, studies \cite{ettinger2020bert} have shown that even these approaches have still struggled with negation. Which is why this study aims to examine the problem in a holistic manner.

\section{Methodology}

This paper addressed two research problems by utilizing Natural Language Processing (NLP) techniques, a lexicon-based approach, and sequence labeling. To guide the design of the aforementioned automated tool, we answered the following two questions:

\begin{itemize}
  \item \textbf{RQ1:} \textit{How do we detect negation in a given text?}
  \item \textbf{RQ2:} \textit{How can a automated system apply negation to appropriate words to improve sentiment analysis?}
\end{itemize}

NLP was used mainly because it guides the extraction of information from texts in their natural form. The most frequent lexicon-based technique for negation is to reverse the polarity of the object that is affected by the negator in a sentence \cite{jurek2015improved}. This study presents an alternative approach by developing a disambiguation function that can be used to average out the polarity scores of the negated word antonyms using five dictionaries. Beyond sentiment detection, we propose using antonyms to construct a human-readable semantic construction of the entire sentence, taking a context-based approach into consideration. First, detecting negation in a sentence requires the use of an automated decontraction system, as some negations may be in contraction mode e.g., \textit{"isn't = \textbf{is not}"}. With the deconstruction pattern shown in algorithm \autoref{alg:decontract}, words that are merged together as contractions are separated so that the system can read the negations appropriately. An annotated dictionary containing negations is created to mirror the words that should be matched when performing tokenized search on a list or text data. The tokenized search is implemented using the list comprehension technique. This method returns a list containing the index of all the negations in a text dataset.

Second, after detecting the negations, the researchers used NLP's Parts of Speech (POS) and Sequential Labeling algorithms to label all of the keywords in the text. With this in place, the machine can readily interpret the POS in each word as to whether the nearest neighbor to a negating signal is a phrase that modifies or qualifies an adjective, verb, or other adverb, or a phrase naming an attribute attached to or grammatically related to a noun to change or describe it. If the POS is an adjective (ADJ) or an adverb (ADV), the system eliminates the negation and inverts the expression. Otherwise, the negation stands.

\subsection{Data Collection}

Our data collection phase was comprised of two stages. The first stage was searching for appropriate datasets that could allow our model to be trained. The complexity in this stage was finding appropriate training datasets that contained negation and, at the same time, were validated by prior literature. The most suitable archive that matched our approach was the Stanford Contradiction Corpora dataset 
. The second stage was the collection of the Stanford Contradiction Corpora dataset from  (\url{https://nlp.stanford.edu/projects/contradiction/}) to test our algorithm. This stage was very critical as it was the basis for demonstrating the efficacy of the algorithm and setting the pace for testing with a larger dataset.

\subsection{Data Pipeline}
\begin{figure}
    \centering
	\includegraphics[clip,width=0.9\linewidth]{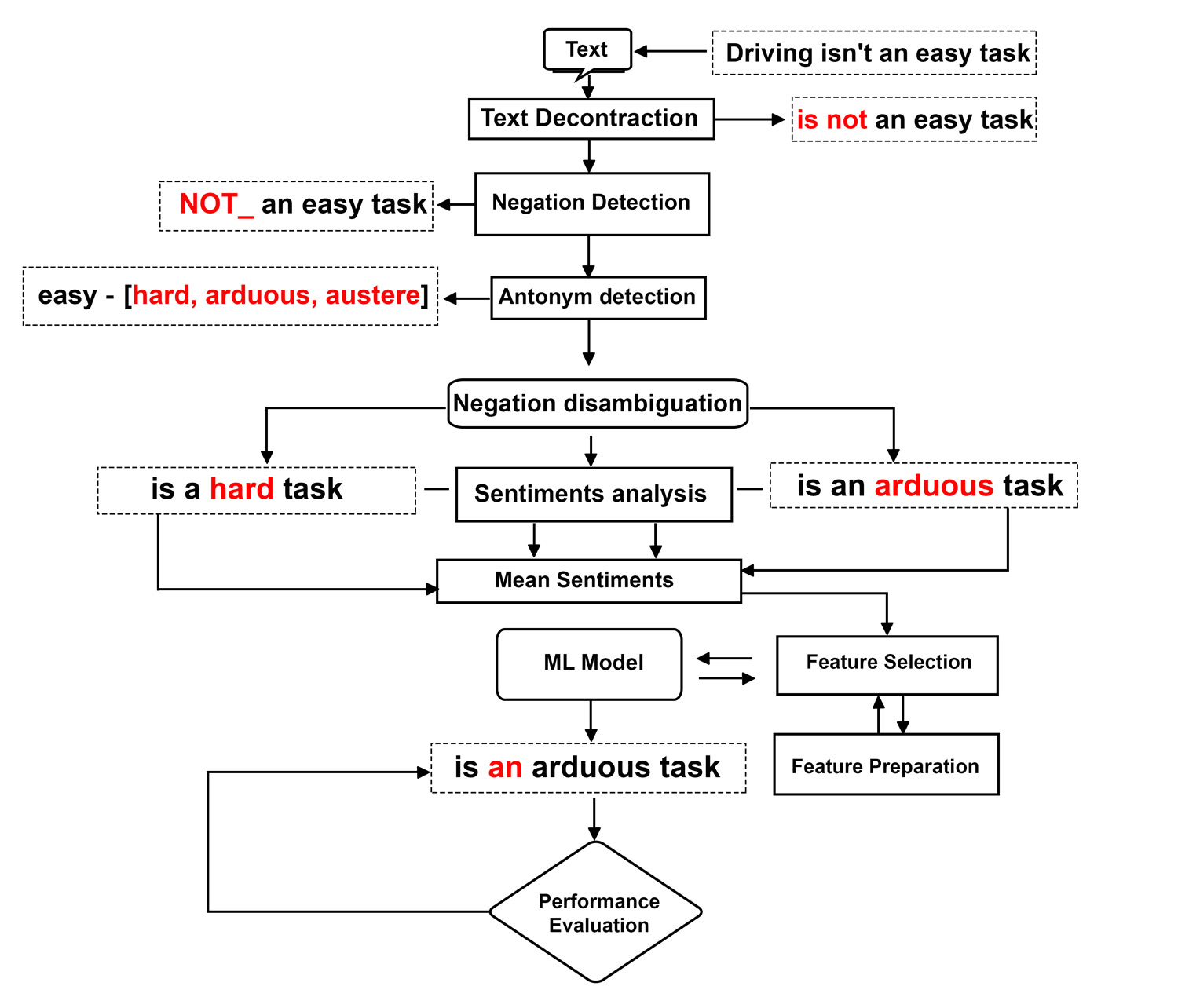}
	\caption{Data Pipeline}
	\label{fig:pipe}
\end{figure}

Considering the nature of the study, our pipelining architecture follows a sequential order in combining various functions down to the disambiguation function as represented in \autoref{fig:pipe}.

\subsubsection{Text Decontraction} 

Words or syllables are often condensed in contractions. By deleting key letters, these shorter versions or contractions of words are generated. In the case of English contractions, one of the vowels is removed from the word. A simple example is when \textit{do not} is shortened to \textit{don't}. The text decontraction function helps in converting each contraction to its enlarged form to ease machine processing. The pseudocode in algorithm \autoref{alg:decontract}, demonstrates a procedure for the decontraction of \textit{n't} and \textit{'t} phrases in a sentence to \textit{not}. This procedure can be implemented in any programming language, but the Python programming language was chosen for this study because of its extensive library for machine learning and natural language processing \cite{muller2016introduction}.

\begin{algorithm}
    \caption{Text Decontraction}\label{alg:decontract}
    \begin{algorithmic}
        \Procedure{decontract}{$text$}
            \If{[n't, 't] IN $text$}
                \State $text\gets [not, not]$ 
                \State return $text$
            \EndIf
        \EndProcedure
    \end{algorithmic}
\end{algorithm}

\subsubsection{Negation Detection} 

Negations, as discussed in the literature review, are not necessarily negative phrases but words that negate the meaning of another. To detect negation, a tokenized search is performed on a list of textual data. An annotated dictionary including negations is constructed to mirror the terms that should be matched. The tokenized search is implemented using the list comprehension approach. This method returns a list with the indexes of all negations in a text data set. The procedure described in algorithm \autoref{alg:neg}, brings negation detection into play. The reference list is first instantiated, displaying the negations in a sentence that should be identified. The procedure then searches through a given text to detect the negation index (key), after the text must have been decontrated using an algorithm \autoref{alg:decontract}.

\begin{algorithm}
    \caption{Detect Negation}\label{alg:neg}
    \begin{algorithmic}
        \Procedure{detect\_neg}{$text$}
            \State $ref\gets ["not", "nor", "never", "neither"]$
            \State $text\gets $text$.split()$
            \For{$ref$ IN $text$}
                \If{(any $item$ IN $ref$)}
                    \State $st$ = set($ref$)
                    \If{($e$ IN $st$)}
                        \For{$i$, $e$ IN enumerate($text$)}
                            \State return $i$ 
                        \EndFor
                    \EndIf
                \EndIf
            \EndFor
        \EndProcedure
    \end{algorithmic}
\end{algorithm}

\subsubsection{Text Antonymization} 

The manual process of antonymizing a given phrase in its natural form involves checking words and their opposites in a dictionary. To do this automatically, we need to make use of some standard dictionaries. In this section, we will be making use of five dictionaries: the Collins Dictionary, the Merriam-Webster Dictionary, synonym.com, thesaurus.com, and NLTK's wordnet. These dictionaries, when combined, provide us with rich content as opposed to using only the popular WordNet database. Furthermore, a Python package called \textit{wordhoard} provides an elaborate library for combining all of these dictionaries \cite{johnbumgarner2021Nov}. Algorithm \autoref{alg:ant} demonstrates the process of antonymization. In some cases, when an antonym is not found, the procedure queries the synonym database, then repeats the process to get the antonym. With this in place, we can easily move ahead to carry out sequence labeling, part of speech tagging, and negation disambiguation based on some pre-defined criteria.

\begin{algorithm}
    \caption{Get Antonymns}\label{alg:ant}
    \begin{algorithmic}
        \State $dictionary[0] \gets collins\_dictionary()$
        \State $dictionary[1] \gets merriam\_webster()$
        \State $dictionary[2] \gets synonym\_com()$
        \State $dictionary[3] \gets thesaurus\_com()$
        \State $dictionary[4] \gets wordnet()$
        
        \State $error\gets not found$
        \Procedure{get\_antonyms}{$word$}
        \State $antonyms$ = Antonyms($word$).find\_antonyms($dict[i]$)
            \If {$error$ NOT IN $antonyms$}
                \State return $antonymns$
            \Else
                \State $synonym \gets get\_synonyms(word)$
                \State $key \gets synonym[0].find\_antonyms(dict[i])$
                \State $antonyms = Antonyms(key)$
                \State return $antonym$ 
                
            \EndIf
        \EndProcedure
    \end{algorithmic}
\end{algorithm}

\par

\subsubsection{Sequence Labeling, Parts of Speech (POS) Tagging, and Negation Disambiguation} 

Sequence labeling in this phase is a form of pattern recognition in machine learning that includes assigning a categorical label to each component of a sequence of observed values using an algorithm (\cite{wiseman2019label}). Part of speech (POS) tagging, which attempts to assign a part of speech to each word in an input sentence or document, is a frequent example of a sequence labeling problem. The POS of a word in a sentence is dependent on its neighbors, thus the use of "sequence" in the term "sequence labeling." The decontract and negation detection procedures play a crucial role in this procedure. Algorithm \autoref{alg:disamb} demonstrates the process of tagging each word in a sentence using the NLTK POS Tagging function. The text input is converted to a list and sequentially labeled using the NLTK library. The label is based on the POS function, which detects distinct parts of speech in text. The tokenized system detects tags associated with \textbf{JJ} (Adjective), \textbf{VBG} (Verb Gerund), and \textbf{VBN} (Verb Past Participle) that have been negated. The main structure of the pseudocode shows that the procedure only removes the negation and inverts the phrase if the POS is an adjective, verb, or adverb. If this is not true, the negation remains in effect. With this in place, the machine can quickly determine whether the neighbor of a negating signal needs to be inverted or not, and it takes three steps towards fulfilling this.

\begin{algorithm}
    \caption{Word Disambiguation}\label{alg:disamb}
    \begin{algorithmic}
        \Procedure{disamb}{$sentence$}
        \State $token\gets nltk.word\_tokenize(sentence)$
        \State $tags\gets nltk.pos\_tag(token)$
            \If {$tags[found]$ = $NEGATED$}
            
                \State $token[int\_index+1]$ = get\_antonyms(token)
                \State $token.pop(int\_index)$
                \State return $token$
            \Else    
                \State return $token$ 
            \EndIf
        \EndProcedure
    \end{algorithmic}
\end{algorithm}

In training a model to understand the patterns for context purposes, we used the BERT model and a recurrent neural network (RNN) to build a dictionary of tags associated with the sequence of words in a sentence. As the antonymization takes place, the neural network model stores the POS of the negated text and an array of the POS of the sentence that accompanies the negation. This value was then plugged into the BERT model and RNN to put the negation into context. The neural network studies the whole POS associated with the sentence that was just analyzed and stores such a pattern in its model. It does not store the exact word because the next sentence that comes might not have the same wording. POS tagging plays a major role in helping the machine learning model learn specific parts of speeches and how to work when similar situations come into play.

\section{Results}

The result from the algorithm was tested with various sentiment analyzers to see if there was a difference in prediction. It is important to point out that most statements that were negated were not necessarily positive or negative statements. A large portion of the dataset consisted of neutral statements. This will also help to test the efficacy of our algorithm because the fact that some statements contain negation does not make the statement a negative sentence, e.g., \textit{"Samuel L. Husk does not work for the Council of Great City Schools."} This statement is neutral, and when returned by our algorithm, it should maintain a neutral stance.

\begin{figure}[H]
    \centering
	\includegraphics[clip,width=1\linewidth]{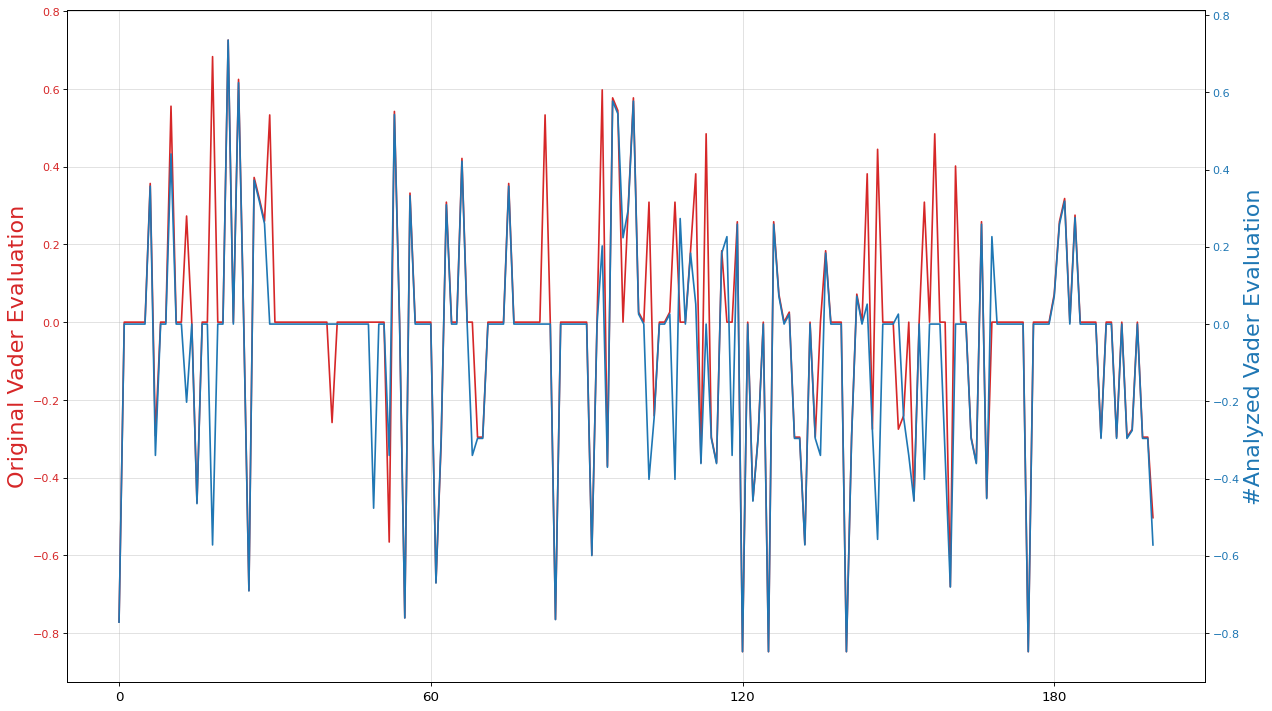}
	\caption{Vader test on the original and antonymized text}
	\label{fig:vader}
\end{figure}

Early results from the Stanford Contradiction Corpora dataset show improvement in predicting sentiments in statements. We will be demonstrating this through the use of three sentiment analyzers. Most importantly, we will be making use of the polarity scores and not the binary output. The result from \autoref{fig:vader} which was derived from a 30\% randomized sample of the original data, shows a slight similarity in output. The randomized sample was used in order not to cluster the figure. The slight similarity in output with \autoref{fig:vader}  was because Vader already implements a system that assigns a negative polarity score to sentences that contain negation. The job of Vader is not to look for a dictionary antonym to use in place of the negated words, but rather to apply a negative score to such a statement. Our algorithm does that by using the exact part of speech tagging against an antonymized word. Vader's improvement is remarkable, but additional improvement is required in terms of using the actual antonyms and not necessarily negating the polarity scores. The other area where Vader played a crucial role was maintaining neutral scores whenever a negation was used instead of going in an extreme direction.

\begin{figure}[!ht]
    \centering
	\includegraphics[clip,width=1\linewidth]{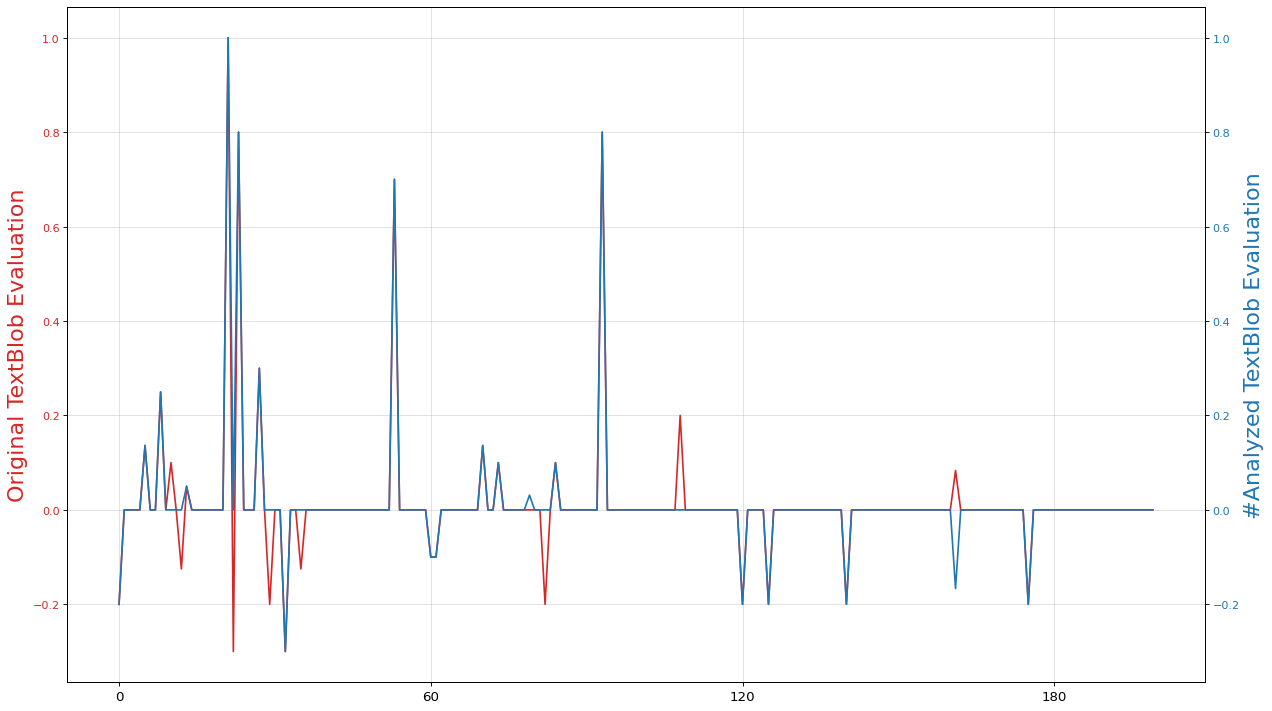}
	\caption{TextBlob test on the original and antonymized text}
	\label{fig:textblob}
\end{figure}

The TextBlob tool works in the same fashion as its Vader counterpart, but with higher accuracy, as we can see from \autoref{fig:textblob}. There is a higher similarity to that of the original text. Though this tool does a great job, the slight difference is very key to making a crucial decision about a delicate situation with data.

\begin{figure}[!ht]
    \centering
	\includegraphics[clip,width=1\linewidth]{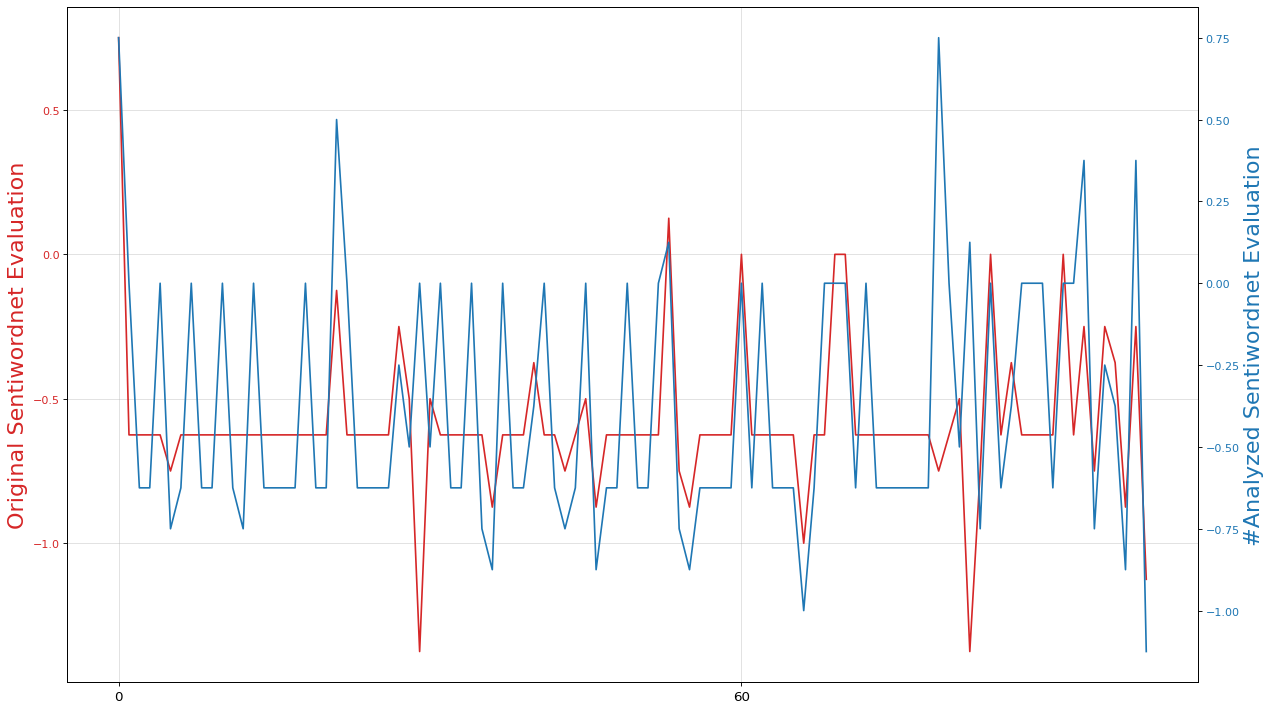}
	\caption{Sentiwordnet test on the original the antonymized text}
	\label{fig:senti}
\end{figure}

The tool that performed the least was the SentiWordNet. This tool needs data to be preprocessed first before passing it through the analyzer, but one key point to note is that the word "not" is a stop word and will end up being eliminated when such analyzers are used. If this is the case, the valuable power derived from such a system will be lost, leaving the user with values that can be misleading. The graph representation of the output is shown in \autoref{fig:senti}.

\begin{figure}[!ht]
    \centering
	\includegraphics[clip,width=0.9\linewidth]{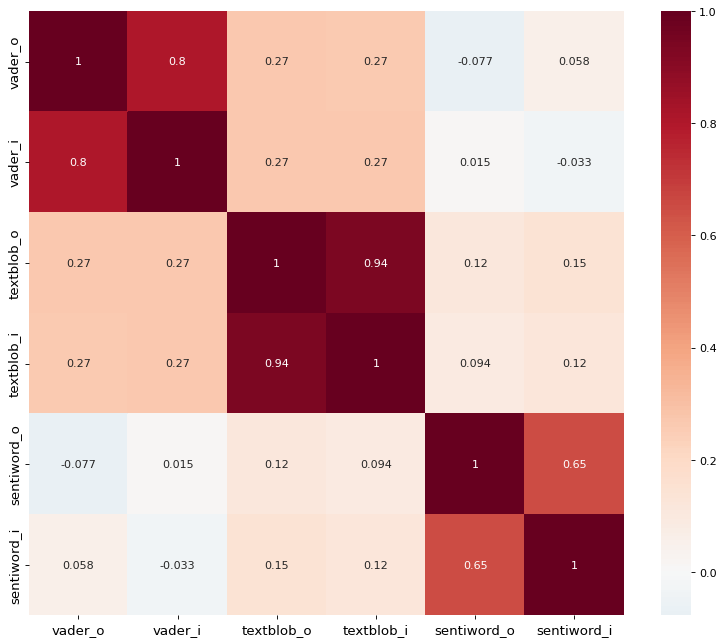}
	\caption{Correlation Heatmap of all the sentiments}
	\label{fig:corre}
\end{figure}

To compare all the outputs together, we make use of a correlogram heatmap shown in \autoref{fig:corre}. The heatmap is presented in the form of a 2D correlation matrix with colored cells and a monochromatic scale \textit{('RdBu\_r')}. The number of measurements that match the dimensional values determines the color value of the cells. This allows us to easily discover patterns of occurrence and spot anomalies. Here is the linear relationship between all the different sentiment analyses. From the matrix, TextBlob gathered a 94\% correlation between the original text and the analyzed text, while Vader returned 80\%. It is also important to note the similarities between the different tools. The TextBlob and Vader output remained unchanged at 27\% while SentiWordNet fluctuated, showing its weaknesses in terms of correlation.

A negation cue is a token that triggers the semantic opposite of a phenomenon in a sentence. As a result, negation connects one statement, \textit{x} to another expression whose meaning is diametrically opposed to \textit{x}'s. This relationship can be expressed in a variety of ways, both syntactically and pragmatically. There are various types of semantic opposition. From both a historical and systematic standpoint, semantic opposition is primarily concerned with negation and opposition in natural language.

Testing our analyzed dataset with the three sentiment analyzers gave a considerable amount of improvement. With a 94\% correlation between the original text and the studied text by Textblob and 80\% correlation with Vader, there was a considerable amount of improvement. The highest impact of our program was on SentiWordNet, which had a correlation score of 65\% showing that our algorithm improved the SentiWordNet output by 35\%.

Furthermore, since the application of artificial neural networks in this research allows for the model to work in a progressive manner, similar to the human mind, the solution shows promise in increasing the accuracy of traditional sentiment analyzers, thereby improving future research in
domains that have to do with sentiment analysis, such as crisis informatics.

\subsection{Limitations}

Negation has been studied as a truth-functional operator, among other things. The precise relationship between negation and words makes negation disambiguation a complex task. Statements caught in the web of negation are also psychologically more complex and difficult to process. This imbalance has been attributed to logic or semantics by several linguists and psychologists, such as the assertion that every negation implies a matching affirmative but not the other way around. This means that there is no special rule for negation disambiguation; rather, the context of the analyzed sentence plays a huge role in understanding which antonym fits for a particular negation signal and not only POS tagging. Since the analysis didn't utilize machine learning for checking grammar, identifying grammar clarity, and checking delivery mistakes, further work needs to be done in this section to improve the neural network that was trained based on POS tagging.

\subsection{Future Work}

Further research is needed to evaluate the system with a larger dataset, especially with data from social media aside from the prepared data that was analyzed in this study. A larger dataset spanning numerous annotations and not necessarily negations can be explored for extensive search and also for improving the neural network algorithm used to store POS tags assigned to analyzed sentences.

The incorporation of a natural grammar-checking program based on deep learning methods will have a significant impact on the analyzed sentences in terms of clarity and error corrections. When dealing with text data, deep learning's capacity to process vast amounts of features makes it incredibly strong. Exploring this technique will improve the work further and, in turn, build a large corpus for automated language understanding and grammar checking.

\section{Conclusion}

This study examined and demonstrated methods for detecting negations in a phrase by examining the distinguishing properties of textual expressions in a sequence and resolving whether or not to invert a word that was negated in a sentence. Three sentiment analyzers were used to test the algorithm, and the results show an improvement in prediction. The improvement was solely because of the way those tools ignored the negation variable, sometimes discarding it as a stop word, resulting in the misclassification of sentences and thereby making accuracy in data prediction with sentiment analyzers lower. This study applied the natural language processing (NLP) approach to find and invert negated terms for proper text classification, resolving the research questions that were asked in the introduction. This approach acts as a lens, reading through a given word sequence utilizing an intelligent NLP algorithm to discover negation signals and invert the keywords that were negated.

Early results suggest that our detailed analysis improved standard sentiment analysis, which sometimes ignores word negations or assigns an inverse polarity score without considering the context of the negated words. The SentiWordNet analyzer was enhanced by 35\%, the Vader analyzer by 20\%, and the TextBlob analyzer was improved by 6\%. Though this study utilized the NLP process, training a machine learning model seemed promising in terms of being able to circumvent the arduous stages necessary in the NLP process. Going by our findings, negation detection and the antonymization of negated words are helpful for researchers involved in crisis evaluation using text analysis. This will allow academics and/or industry practitioners who derive insights from textual data to conduct their analyses with greater accuracy than was previously possible. 


\bibliographystyle{ACM-Reference-Format}
\bibliography{acmart}


\begin{thebibliography}{38}


\ifx \showCODEN    \undefined \def \showCODEN     #1{\unskip}     \fi
\ifx \showDOI      \undefined \def \showDOI       #1{#1}\fi
\ifx \showISBNx    \undefined \def \showISBNx     #1{\unskip}     \fi
\ifx \showISBNxiii \undefined \def \showISBNxiii  #1{\unskip}     \fi
\ifx \showISSN     \undefined \def \showISSN      #1{\unskip}     \fi
\ifx \showLCCN     \undefined \def \showLCCN      #1{\unskip}     \fi
\ifx \shownote     \undefined \def \shownote      #1{#1}          \fi
\ifx \showarticletitle \undefined \def \showarticletitle #1{#1}   \fi
\ifx \showURL      \undefined \def \showURL       {\relax}        \fi
\providecommand\bibfield[2]{#2}
\providecommand\bibinfo[2]{#2}
\providecommand\natexlab[1]{#1}
\providecommand\showeprint[2][]{arXiv:#2}

\bibitem[Abirami and Gayathri(2017)]%
        {abirami2017survey}
\bibfield{author}{\bibinfo{person}{AM Abirami} {and} \bibinfo{person}{V
  Gayathri}.} \bibinfo{year}{2017}\natexlab{}.
\newblock \showarticletitle{A survey on sentiment analysis methods and
  approach}. In \bibinfo{booktitle}{\emph{2016 Eighth International Conference
  on Advanced Computing (ICoAC)}}. IEEE, \bibinfo{pages}{72--76}.
\newblock


\bibitem[Agarwal and Yu(2010)]%
        {agarwal2010biomedical}
\bibfield{author}{\bibinfo{person}{Shashank Agarwal} {and}
  \bibinfo{person}{Hong Yu}.} \bibinfo{year}{2010}\natexlab{}.
\newblock \showarticletitle{Biomedical negation scope detection with
  conditional random fields}.
\newblock \bibinfo{journal}{\emph{Journal of the American medical informatics
  association}} \bibinfo{volume}{17}, \bibinfo{number}{6}
  (\bibinfo{year}{2010}), \bibinfo{pages}{696--701}.
\newblock


\bibitem[Baccianella et~al\mbox{.}(2010)]%
        {baccianella2010sentiwordnet}
\bibfield{author}{\bibinfo{person}{Stefano Baccianella},
  \bibinfo{person}{Andrea Esuli}, {and} \bibinfo{person}{Fabrizio Sebastiani}.}
  \bibinfo{year}{2010}\natexlab{}.
\newblock \showarticletitle{Sentiwordnet 3.0: An enhanced lexical resource for
  sentiment analysis and opinion mining}. In
  \bibinfo{booktitle}{\emph{Proceedings of the Seventh International Conference
  on Language Resources and Evaluation (LREC'10)}}.
\newblock


\bibitem[Baker(1969)]%
        {baker1969double}
\bibfield{author}{\bibinfo{person}{C~Leroy Baker}.}
  \bibinfo{year}{1969}\natexlab{}.
\newblock \showarticletitle{Double negatives}.
\newblock \bibinfo{journal}{\emph{Research on Language \& Social Interaction}}
  \bibinfo{volume}{1}, \bibinfo{number}{1} (\bibinfo{year}{1969}),
  \bibinfo{pages}{16--40}.
\newblock


\bibitem[Chapman et~al\mbox{.}(2001)]%
        {chapman2001simple}
\bibfield{author}{\bibinfo{person}{Wendy~W Chapman}, \bibinfo{person}{Will
  Bridewell}, \bibinfo{person}{Paul Hanbury}, \bibinfo{person}{Gregory~F
  Cooper}, {and} \bibinfo{person}{Bruce~G Buchanan}.}
  \bibinfo{year}{2001}\natexlab{}.
\newblock \showarticletitle{A simple algorithm for identifying negated findings
  and diseases in discharge summaries}.
\newblock \bibinfo{journal}{\emph{Journal of biomedical informatics}}
  \bibinfo{volume}{34}, \bibinfo{number}{5} (\bibinfo{year}{2001}),
  \bibinfo{pages}{301--310}.
\newblock


\bibitem[Coombs(2015)]%
        {coombs2015value}
\bibfield{author}{\bibinfo{person}{W~Timothy Coombs}.}
  \bibinfo{year}{2015}\natexlab{}.
\newblock \showarticletitle{The value of communication during a crisis:
  Insights from strategic communication research}.
\newblock \bibinfo{journal}{\emph{Business horizons}} \bibinfo{volume}{58},
  \bibinfo{number}{2} (\bibinfo{year}{2015}), \bibinfo{pages}{141--148}.
\newblock


\bibitem[Devlin et~al\mbox{.}(2018)]%
        {devlin2018bert}
\bibfield{author}{\bibinfo{person}{Jacob Devlin}, \bibinfo{person}{Ming-Wei
  Chang}, \bibinfo{person}{Kenton Lee}, {and} \bibinfo{person}{Kristina
  Toutanova}.} \bibinfo{year}{2018}\natexlab{}.
\newblock \showarticletitle{Bert: Pre-training of deep bidirectional
  transformers for language understanding}.
\newblock \bibinfo{journal}{\emph{arXiv preprint arXiv:1810.04805}}
  (\bibinfo{year}{2018}).
\newblock


\bibitem[Ettinger(2020)]%
        {ettinger2020bert}
\bibfield{author}{\bibinfo{person}{Allyson Ettinger}.}
  \bibinfo{year}{2020}\natexlab{}.
\newblock \showarticletitle{What BERT is not: Lessons from a new suite of
  psycholinguistic diagnostics for language models}.
\newblock \bibinfo{journal}{\emph{Transactions of the Association for
  Computational Linguistics}}  \bibinfo{volume}{8} (\bibinfo{year}{2020}),
  \bibinfo{pages}{34--48}.
\newblock


\bibitem[Fu et~al\mbox{.}(2018)]%
        {fu2018lexicon}
\bibfield{author}{\bibinfo{person}{Xianghua Fu}, \bibinfo{person}{Jingying
  Yang}, \bibinfo{person}{Jianqiang Li}, \bibinfo{person}{Min Fang}, {and}
  \bibinfo{person}{Huihui Wang}.} \bibinfo{year}{2018}\natexlab{}.
\newblock \showarticletitle{Lexicon-enhanced LSTM with attention for general
  sentiment analysis}.
\newblock \bibinfo{journal}{\emph{IEEE Access}}  \bibinfo{volume}{6}
  (\bibinfo{year}{2018}), \bibinfo{pages}{71884--71891}.
\newblock


\bibitem[Gardner et~al\mbox{.}(2018)]%
        {gardner2018allennlp}
\bibfield{author}{\bibinfo{person}{Matt Gardner}, \bibinfo{person}{Joel Grus},
  \bibinfo{person}{Mark Neumann}, \bibinfo{person}{Oyvind Tafjord},
  \bibinfo{person}{Pradeep Dasigi}, \bibinfo{person}{Nelson Liu},
  \bibinfo{person}{Matthew Peters}, \bibinfo{person}{Michael Schmitz}, {and}
  \bibinfo{person}{Luke Zettlemoyer}.} \bibinfo{year}{2018}\natexlab{}.
\newblock \showarticletitle{Allennlp: A deep semantic natural language
  processing platform}.
\newblock \bibinfo{journal}{\emph{arXiv preprint arXiv:1803.07640}}
  (\bibinfo{year}{2018}).
\newblock


\bibitem[Horn and Wansing(2020)]%
        {sepnegation}
\bibfield{author}{\bibinfo{person}{Laurence~R. Horn} {and}
  \bibinfo{person}{Heinrich Wansing}.} \bibinfo{year}{2020}\natexlab{}.
\newblock \showarticletitle{{Negation}}.
\newblock In \bibinfo{booktitle}{\emph{The {Stanford} Encyclopedia of
  Philosophy} (\bibinfo{edition}{{S}pring 2020} ed.)},
  \bibfield{editor}{\bibinfo{person}{Edward~N. Zalta}} (Ed.).
  \bibinfo{publisher}{Metaphysics Research Lab, Stanford University}.
\newblock


\bibitem[Hutto and Gilbert(2014)]%
        {hutto2014vader}
\bibfield{author}{\bibinfo{person}{Clayton Hutto} {and} \bibinfo{person}{Eric
  Gilbert}.} \bibinfo{year}{2014}\natexlab{}.
\newblock \showarticletitle{Vader: A parsimonious rule-based model for
  sentiment analysis of social media text}. In
  \bibinfo{booktitle}{\emph{Proceedings of the international AAAI conference on
  web and social media}}, Vol.~\bibinfo{volume}{8}. \bibinfo{pages}{216--225}.
\newblock


\bibitem[Ifeanyi et~al\mbox{.}(2014)]%
        {ifeanyi2014text}
\bibfield{author}{\bibinfo{person}{Nwakanma Ifeanyi}, \bibinfo{person}{Oluigbo
  Ikenna}, {and} \bibinfo{person}{Okpala Izunna}.}
  \bibinfo{year}{2014}\natexlab{}.
\newblock \showarticletitle{Text--To--Speech Synthesis (TTS)}.
\newblock \bibinfo{journal}{\emph{International Journal of Research in
  Information Technology (IJRIT)}} \bibinfo{volume}{2}, \bibinfo{number}{5}
  (\bibinfo{year}{2014}), \bibinfo{pages}{154--163}.
\newblock


\bibitem[Izunna et~al\mbox{.}(2022)]%
        {okpala2022perception}
\bibfield{author}{\bibinfo{person}{Okpala Izunna}, \bibinfo{person}{Rodriguez
  Guillermo, Romera}, \bibinfo{person}{Zheng Weibing}, \bibinfo{person}{Halse
  Shane}, {and} \bibinfo{person}{Kropczynski Jess}.}
  \bibinfo{year}{2022}\natexlab{}.
\newblock \showarticletitle{Perception Analysis: Pro- and Anti- Vaccine
  Classification with NLP and Machine Learning}.
\newblock \bibinfo{journal}{\emph{Proceedings of the 55th Hawaii International
  Conference on System Sciences}} (\bibinfo{year}{2022}),
  \bibinfo{pages}{2981--2990}.
\newblock


\bibitem[Jiao and Qu(2019)]%
        {jiao2019proposal}
\bibfield{author}{\bibinfo{person}{Yiru Jiao} {and} \bibinfo{person}{Qing-Xing
  Qu}.} \bibinfo{year}{2019}\natexlab{}.
\newblock \showarticletitle{A proposal for Kansei knowledge extraction method
  based on natural language processing technology and online product reviews}.
\newblock \bibinfo{journal}{\emph{Computers in Industry}}
  \bibinfo{volume}{108} (\bibinfo{year}{2019}), \bibinfo{pages}{1--11}.
\newblock


\bibitem[Jiaxuan(2010)]%
        {jiaxuan2010division}
\bibfield{author}{\bibinfo{person}{SHEN Jiaxuan}.}
  \bibinfo{year}{2010}\natexlab{}.
\newblock \showarticletitle{Division of negatives and noun/verb division in
  English and Chinese [J]}.
\newblock \bibinfo{journal}{\emph{Studies of the Chinese Language}}
  \bibinfo{volume}{5} (\bibinfo{year}{2010}).
\newblock


\bibitem[johnbumgarner(2021)]%
        {johnbumgarner2021Nov}
\bibfield{author}{\bibinfo{person}{johnbumgarner}.}
  \bibinfo{year}{2021}\natexlab{}.
\newblock \bibinfo{title}{{wordhoard}}.
\newblock
\newblock
\urldef\tempurl%
\url{https://github.com/johnbumgarner/wordhoard}
\showURL{%
\tempurl}
\newblock
\shownote{[Online; accessed 30. Nov. 2021]}.


\bibitem[Jurek et~al\mbox{.}(2015)]%
        {jurek2015improved}
\bibfield{author}{\bibinfo{person}{Anna Jurek}, \bibinfo{person}{Maurice~D
  Mulvenna}, {and} \bibinfo{person}{Yaxin Bi}.}
  \bibinfo{year}{2015}\natexlab{}.
\newblock \showarticletitle{Improved lexicon-based sentiment analysis for
  social media analytics}.
\newblock \bibinfo{journal}{\emph{Security Informatics}} \bibinfo{volume}{4},
  \bibinfo{number}{1} (\bibinfo{year}{2015}), \bibinfo{pages}{1--13}.
\newblock


\bibitem[Khandelwal and Sawant(2019)]%
        {khandelwal2019negbert}
\bibfield{author}{\bibinfo{person}{Aditya Khandelwal} {and}
  \bibinfo{person}{Suraj Sawant}.} \bibinfo{year}{2019}\natexlab{}.
\newblock \showarticletitle{Negbert: A transfer learning approach for negation
  detection and scope resolution}.
\newblock \bibinfo{journal}{\emph{arXiv preprint arXiv:1911.04211}}
  (\bibinfo{year}{2019}).
\newblock


\bibitem[Kimberly(2021)]%
        {kimberly2021double}
\bibfield{author}{\bibinfo{person}{Joki Kimberly}.}
  \bibinfo{year}{2021}\natexlab{}.
\newblock \bibinfo{title}{Double Negatives: 3 Rules You Must Know}.
\newblock
\newblock
\urldef\tempurl%
\url{https://www.grammarly.com/blog/3-things-you-must-know-about-double-negatives}
\showURL{%
\tempurl}
\newblock
\shownote{[Online; accessed 23. Jul. 2021]}.


\bibitem[Kovaleva et~al\mbox{.}(2021)]%
        {kovaleva2021bert}
\bibfield{author}{\bibinfo{person}{Olga Kovaleva}, \bibinfo{person}{Saurabh
  Kulshreshtha}, \bibinfo{person}{Anna Rogers}, {and} \bibinfo{person}{Anna
  Rumshisky}.} \bibinfo{year}{2021}\natexlab{}.
\newblock \showarticletitle{BERT Busters: Outlier Dimensions that Disrupt
  Transformers}.
\newblock \bibinfo{journal}{\emph{arXiv preprint arXiv:2105.06990}}
  (\bibinfo{year}{2021}).
\newblock


\bibitem[Liang(2020)]%
        {Liang2020May}
\bibfield{author}{\bibinfo{person}{Xu Liang}.} \bibinfo{year}{2020}\natexlab{}.
\newblock \showarticletitle{{Treat Negation Stopwords Differently According to
  Your NLP Task}}.
\newblock \bibinfo{journal}{\emph{Medium}} (\bibinfo{date}{May}
  \bibinfo{year}{2020}).
\newblock
\urldef\tempurl%
\url{https://towardsdatascience.com/treat-negation-stopwords-differently-according-to-your-nlp-task-e5a59ab7c91f}
\showURL{%
\tempurl}


\bibitem[Manzoor et~al\mbox{.}(2019)]%
        {manzoor2019fake}
\bibfield{author}{\bibinfo{person}{Syed~Ishfaq Manzoor}, \bibinfo{person}{Jimmy
  Singla}, {et~al\mbox{.}}} \bibinfo{year}{2019}\natexlab{}.
\newblock \showarticletitle{Fake news detection using machine learning
  approaches: A systematic review}. In \bibinfo{booktitle}{\emph{2019 3rd
  International Conference on Trends in Electronics and Informatics (ICOEI)}}.
  IEEE, \bibinfo{pages}{230--234}.
\newblock


\bibitem[M{\"u}ller and Guido(2016)]%
        {muller2016introduction}
\bibfield{author}{\bibinfo{person}{Andreas~C M{\"u}ller} {and}
  \bibinfo{person}{Sarah Guido}.} \bibinfo{year}{2016}\natexlab{}.
\newblock \bibinfo{booktitle}{\emph{Introduction to machine learning with
  Python: a guide for data scientists}}.
\newblock \bibinfo{publisher}{" O'Reilly Media, Inc."}.
\newblock


\bibitem[Mutalik et~al\mbox{.}(2001)]%
        {mutalik2001use}
\bibfield{author}{\bibinfo{person}{Pradeep~G Mutalik},
  \bibinfo{person}{Aniruddha Deshpande}, {and} \bibinfo{person}{Prakash~M
  Nadkarni}.} \bibinfo{year}{2001}\natexlab{}.
\newblock \showarticletitle{Use of general-purpose negation detection to
  augment concept indexing of medical documents: a quantitative study using the
  UMLS}.
\newblock \bibinfo{journal}{\emph{Journal of the American Medical Informatics
  Association}} \bibinfo{volume}{8}, \bibinfo{number}{6}
  (\bibinfo{year}{2001}), \bibinfo{pages}{598--609}.
\newblock


\bibitem[Ovalle and Guerzoni(2004)]%
        {ovalle2004double}
\bibfield{author}{\bibinfo{person}{Luis~Alonso Ovalle} {and}
  \bibinfo{person}{Elena Guerzoni}.} \bibinfo{year}{2004}\natexlab{}.
\newblock \showarticletitle{Double negatives, negative concord and
  metalinguistic negation}. In \bibinfo{booktitle}{\emph{CLS 38.1: The main
  session. Proceedings from the main session of the 38th meeting of the Chicago
  Linguistic Society}}. Citeseer, \bibinfo{pages}{15--31}.
\newblock


\bibitem[Phadke and Devane(2017)]%
        {phadke2017multilingual}
\bibfield{author}{\bibinfo{person}{Madhura~Mandar Phadke} {and}
  \bibinfo{person}{Satish~R Devane}.} \bibinfo{year}{2017}\natexlab{}.
\newblock \showarticletitle{Multilingual Machine translation: An analytical
  study}. In \bibinfo{booktitle}{\emph{2017 International Conference on
  Intelligent Computing and Control Systems (ICICCS)}}. IEEE,
  \bibinfo{pages}{881--884}.
\newblock


\bibitem[Quilty(2019)]%
        {quilty2019university}
\bibfield{author}{\bibinfo{person}{Ellen Quilty}.}
  \bibinfo{year}{2019}\natexlab{}.
\newblock \showarticletitle{University Library: AMA Referencing (Vancouver):
  Internet \& Social Media}.
\newblock  (\bibinfo{year}{2019}).
\newblock


\bibitem[R{\'\i}o~Zamora et~al\mbox{.}(2015)]%
        {rio2015comparative}
\bibfield{author}{\bibinfo{person}{Vanesa~del R{\'\i}o~Zamora} {et~al\mbox{.}}}
  \bibinfo{year}{2015}\natexlab{}.
\newblock \showarticletitle{Comparative Study of the Use of Double Negatives by
  Native English Speakers and Spanish Learners of English}.
\newblock  (\bibinfo{year}{2015}).
\newblock


\bibitem[Rokach et~al\mbox{.}(2008)]%
        {rokach2008negation}
\bibfield{author}{\bibinfo{person}{Lior Rokach}, \bibinfo{person}{Roni Romano},
  {and} \bibinfo{person}{Oded Maimon}.} \bibinfo{year}{2008}\natexlab{}.
\newblock \showarticletitle{Negation recognition in medical narrative reports}.
\newblock \bibinfo{journal}{\emph{Information Retrieval}} \bibinfo{volume}{11},
  \bibinfo{number}{6} (\bibinfo{year}{2008}), \bibinfo{pages}{499--538}.
\newblock


\bibitem[Tixier et~al\mbox{.}(2016)]%
        {tixier2016automated}
\bibfield{author}{\bibinfo{person}{Antoine J-P Tixier},
  \bibinfo{person}{Matthew~R Hallowell}, \bibinfo{person}{Balaji Rajagopalan},
  {and} \bibinfo{person}{Dean Bowman}.} \bibinfo{year}{2016}\natexlab{}.
\newblock \showarticletitle{Automated content analysis for construction safety:
  A natural language processing system to extract precursors and outcomes from
  unstructured injury reports}.
\newblock \bibinfo{journal}{\emph{Automation in Construction}}
  \bibinfo{volume}{62} (\bibinfo{year}{2016}), \bibinfo{pages}{45--56}.
\newblock


\bibitem[Topal et~al\mbox{.}(2021)]%
        {topal2021exploring}
\bibfield{author}{\bibinfo{person}{M~Onat Topal}, \bibinfo{person}{Anil Bas},
  {and} \bibinfo{person}{Imke van Heerden}.} \bibinfo{year}{2021}\natexlab{}.
\newblock \showarticletitle{Exploring transformers in natural language
  generation: Gpt, bert, and xlnet}.
\newblock \bibinfo{journal}{\emph{arXiv preprint arXiv:2102.08036}}
  (\bibinfo{year}{2021}).
\newblock


\bibitem[Udebuana et~al\mbox{.}(2019)]%
        {udebuana2019analysis}
\bibfield{author}{\bibinfo{person}{Okpala~Izunna Udebuana},
  \bibinfo{person}{Ijioma~Patricia Ngozi}, {and}
  \bibinfo{person}{Emejulu~Augustine Obiajulu}.}
  \bibinfo{year}{2019}\natexlab{}.
\newblock \showarticletitle{Analysis of Evaluated Sentiments; a
  Pseudo-Linguistic Approach and Online Acceptability Index for Decision-Making
  with Data: Nigerian Election in View}.
\newblock \bibinfo{journal}{\emph{Computing}} \bibinfo{volume}{7},
  \bibinfo{number}{2} (\bibinfo{year}{2019}), \bibinfo{pages}{39--44}.
\newblock


\bibitem[Verma et~al\mbox{.}(2011)]%
        {verma2011natural}
\bibfield{author}{\bibinfo{person}{Sudha Verma}, \bibinfo{person}{Sarah
  Vieweg}, \bibinfo{person}{William Corvey}, \bibinfo{person}{Leysia Palen},
  \bibinfo{person}{James Martin}, \bibinfo{person}{Martha Palmer},
  \bibinfo{person}{Aaron Schram}, {and} \bibinfo{person}{Kenneth Anderson}.}
  \bibinfo{year}{2011}\natexlab{}.
\newblock \showarticletitle{Natural language processing to the rescue?
  extracting" situational awareness" tweets during mass emergency}. In
  \bibinfo{booktitle}{\emph{Proceedings of the International AAAI Conference on
  Web and Social Media}}, Vol.~\bibinfo{volume}{5}.
\newblock


\bibitem[Webster and Kit(1992)]%
        {webster1992tokenization}
\bibfield{author}{\bibinfo{person}{Jonathan~J Webster} {and}
  \bibinfo{person}{Chunyu Kit}.} \bibinfo{year}{1992}\natexlab{}.
\newblock \showarticletitle{Tokenization as the initial phase in NLP}. In
  \bibinfo{booktitle}{\emph{COLING 1992 Volume 4: The 14th International
  Conference on Computational Linguistics}}.
\newblock


\bibitem[Widdows(2003)]%
        {widdows2003orthogonal}
\bibfield{author}{\bibinfo{person}{Dominic Widdows}.}
  \bibinfo{year}{2003}\natexlab{}.
\newblock \showarticletitle{Orthogonal negation in vector spaces for modelling
  word-meanings and document retrieval}. In
  \bibinfo{booktitle}{\emph{Proceedings of the 41st annual meeting of the
  association for computational linguistics}}. \bibinfo{pages}{136--143}.
\newblock


\bibitem[Wiseman and Stratos(2019)]%
        {wiseman2019label}
\bibfield{author}{\bibinfo{person}{Sam Wiseman} {and} \bibinfo{person}{Karl
  Stratos}.} \bibinfo{year}{2019}\natexlab{}.
\newblock \showarticletitle{Label-agnostic sequence labeling by copying nearest
  neighbors}.
\newblock \bibinfo{journal}{\emph{arXiv preprint arXiv:1906.04225}}
  (\bibinfo{year}{2019}).
\newblock


\bibitem[Yu et~al\mbox{.}(2017)]%
        {yu2017refining}
\bibfield{author}{\bibinfo{person}{Liang-Chih Yu}, \bibinfo{person}{Jin Wang},
  \bibinfo{person}{K~Robert Lai}, {and} \bibinfo{person}{Xuejie Zhang}.}
  \bibinfo{year}{2017}\natexlab{}.
\newblock \showarticletitle{Refining word embeddings using intensity scores for
  sentiment analysis}.
\newblock \bibinfo{journal}{\emph{IEEE/ACM Transactions on Audio, Speech, and
  Language Processing}} \bibinfo{volume}{26}, \bibinfo{number}{3}
  (\bibinfo{year}{2017}), \bibinfo{pages}{671--681}.
\newblock


\end{thebibliography}


\end{document}